\documentclass[a4paper]{article}

\usepackage{INTERSPEECH2020}
\usepackage{xcolor}
\usepackage{subcaption}
\usepackage{bbold}

\newcommand{\x}{\mathbf{x}}
\newcommand{\y}{\mathbf{y}}

\title{Predicting detection filters for small footprint \\ open-vocabulary keyword spotting}
\name{Th\'{e}odore Bluche, Thibault Gisselbrecht}
\address{
  Sonos Inc., Paris, France}
\email{first.last@sonos.com}

\begin{document}

\maketitle
\begin{abstract}
In this paper, we propose a fully-neural approach to open-vocabulary keyword spotting, 
that allows the users to include a customizable voice interface to their device and that does not require task-specific data. We present a keyword detection neural network weighing less than 250KB, in which the topmost layer performing keyword detection is predicted by an auxiliary network, that may be run offline to generate a detector for any keyword. We show that the proposed model outperforms acoustic keyword spotting baselines by a large margin on two tasks of detecting keywords in utterances and three tasks of detecting isolated speech commands. We also propose a method to fine-tune the model when specific training data is available for some keywords, which yields a performance similar to a standard speech command neural network while keeping the ability of the model to be applied to new keywords.
\end{abstract}

\section{Introduction}
\label{sec:intro}
The recent advances in automatic speech recognition (ASR), reaching close to human recognition performance~\cite{xiong2016achieving}, paved the way to natural language interaction in everyday life, making voice become a natural interface for the communication with objects. 
Such large-vocabulary ASR systems demand a lot of resources and computing power, but it has been shown~\cite{saade2018spoken} that the voice interface can run on device when the tasks are known, in a closed-ontology setting. For many interactions it may even be reduced to the detection of specific keywords, allowing to build systems small enough to run on micro-controllers~(MCUs), which are cheap and have a low energy consumption~\cite{helloedge}.
%

A significant amount of tiny neural networks for keyword spotting~(KWS) have been proposed in the past few years~\cite{sainath2015convolutional,coucke2019efficient}. These models yield very good keyword classification results and can run on MCUs. However, they require specific training data containing the keywords to be detected at inference. They lack flexibility because data should be collected and a new model should be trained every time a new keyword is added. On the other hand, traditional KWS approaches are either based on the output of an ASR system, looking for keywords in the transcript~\cite{garofolo2000trec} or in the word~\cite{mamou2006spoken,miller2007rapid} or phone~\cite{brown1997open,chen2017confidence} lattice, or based on acoustic models of phones, allowing to build models for keywords and ``background" and computing likelihood ratios between the two~\cite{szoke2005comparison}.
End-to-end acoustic models predicting the character or phone sequence directly lead to efficient decoding and keyword spotting~\cite{hwang2015online,lengerich2016end,chen2018sequence,bluche2020} and can take into account the confusions of the network to improve the keyword models~\cite{lugosch2018donut,yang2018automatic}. 
These methods do not need keyword-specific training data, but still require some post-processing and a non-trivial confidence score calibration to transform the frame-wise phone scores into keyword confidences.

Recently, ASR-free approaches have been proposed, which consist in computing embeddings for both the audio and the keyword pronunciation and directly predict a keyword detection score~\cite{audhkhasi2017end,sacchi2019open}. They combine the simplicity of end-to-end KWS methods and the flexibility of acoustic KWS, and do not require specific training data.
In~\cite{audhkhasi2017end}, the whole spoken utterance is embedded into a single vector with a recurrent auto-encoder. Similarly, the keyword is embedded into a vector using an auto-encoder of the phone sequence. The concatenation of both vectors is fed to a small neural network predicting whether the keyword appears in the utterance. In~\cite{sacchi2019open}, different recurrent neural networks are trained to predict the word and phone embeddings. The classification is based on the distance between the keyword and utterance embedding. 

In a similar vein, we propose a fully neural architecture for KWS which can be trained on generic ASR data, without specific examples of the keywords to be detected at inference. It is made of three components. An acoustic encoder, composed of a stack of recurrent layers, is pretrained as a quantized ASR acoustic model. Its intermediate features are fed to a convolutional keyword detector network trained to output keyword confidences. The weights of the latter are predicted by a keyword encoder neural network: a recurrent neural network applied to the keyword pronunciations to predict the weights of the topmost convolution kernel of the keyword detector network. This idea is similar to other works on dynamic convolution filters in computer vision for weather prediction~\cite{klein2015dynamic}, visual question answering~\cite{noh2016image}, or video and stereo prediction~\cite{jia2016dynamic}.

We experimented this approach on two tasks: a continuous KWS task where keywords are detected inside queries formulated in natural language, and a speech command task where the goal is to predict one command among a predefined set. We compare this system to acoustic KWS approaches and we show that the proposed neural approach outperforms them by a large margin. We also show how the model may be fine-tuned with specific training data to get close to the performance of an end-to-end KWS classification model, without losing the ability of the model to detect new out-of-vocabulary keywords.

The remaining of the paper is divided as follows. 
The proposed model is described in Section~\ref{sec:model}. 
We report the experimental results on the two tasks in Section~\ref{sec:results} 
and conclude the paper in Section~\ref{sec:conclu}.

\section{Keyword spotting neural network}
\label{sec:model}

When keywords to detect are known in advance, and when training data containing those keywords are available, a neural network can be trained in an end-to-end fashion to detect them~\cite{sainath2015convolutional,coucke2019efficient}. In this paper, we present a method to create such a neural network for any arbitrary keyword defined post-training, which does not require training data specific to these keywords.

\subsection{Keyword spotting neural network}

The neural network is made of a stack of unidirectional LSTM layers, followed by two convolutional layers. It has one sigmoid output for each keyword, and is trained on a generic speech dataset. 
The top convolutional layer of the neural network computes the probability of detecting each keyword at each timestep. For keyword $k$, the output sequence $\y_k$ is computed as follows
\begin{equation}
    \y_{k} = y_{k,1} \ldots y_{k,T} = \sigma ( \theta_k \ast F(\x; \theta_F) )\label{eq:kw-classif},
\end{equation}
where 
$\sigma$ is the sigmoid function, $\ast$ is the convolution operation,
$F(\x; \theta_F)$ represents the lower layers of the neural network with parameters $\theta_F$ applied to input $\x$,
and $\theta_k$ is the convolution kernel corresponding to keyword $k$.

\begin{figure}[!htb]
\centering
  \includegraphics[width=0.8\linewidth]{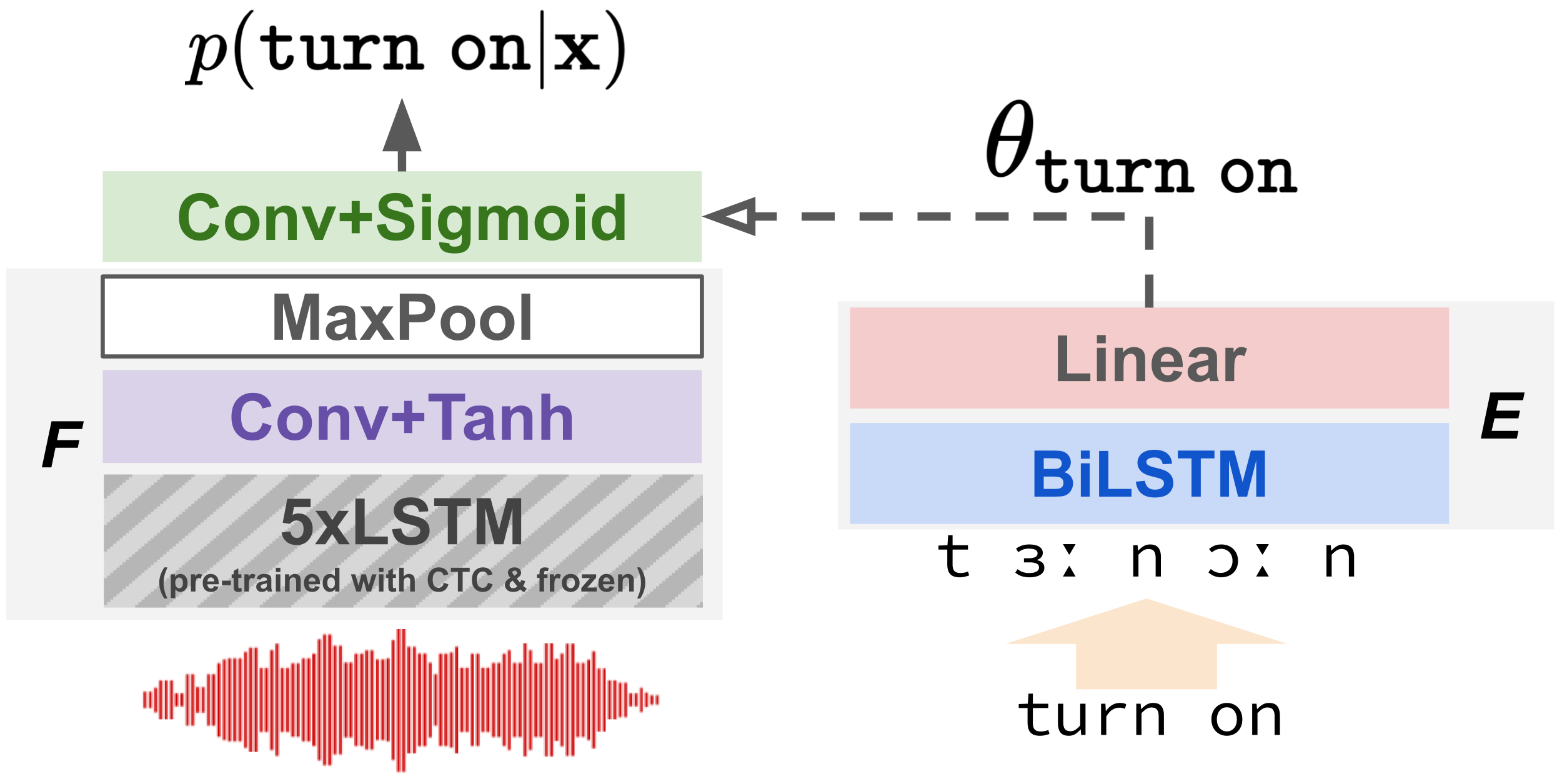}
\caption{Proposed model. A keyword encoder (right) predicts the weights of the top convolutional filter of the keyword spotting network (left), used to detect the keyword.}
\label{fig:nf-model}
\end{figure}

Since the keywords to be detected are not known during the training phase, and because the network is an open-vocabulary KWS model, the parameters $\theta_k$ of the top layer cannot be trained directly. They are predicted by an auxiliary neural network: a keyword encoder, as shown in Figure~\ref{fig:nf-model}.

The keyword encoder $E$ is a neural network applied to the phone sequence $\pi_k$ of a keyword $k$. It outputs the parameters $\theta_k$ used to detect the keyword with the KWS model:
\begin{equation}
    \theta_k = E(\pi_k; \theta_E),\label{eq:kw-encoder}
\end{equation}
where $\theta_E$ represents the parameters of the keyword encoder network. In this work, the keyword encoder is a bidirectional LSTM network, followed by an affine transform. 

\subsection{Inference}

At inference, the network is first configured to detect a set of $n$ keywords $\mathcal{K} = \{k_1, \ldots, k_n\}$. For each keyword $k \in \mathcal{K}$, the phone sequence $\pi_k$ is retrieved from a pronunciation lexicon or a grapheme-to-phoneme converter, and 
 convolution kernel $\theta_k$ is computed by the keyword encoder (eq.~\ref{eq:kw-encoder}). 
 Then, the top convolution layer can be created with the set of computed kernels $\{\theta_k~;~k \in \mathcal{K}\}$, and the KWS network is ready to be applied to the input audio (eq.~\ref{eq:kw-classif}).

\subsection{Training}
\label{sec:training}
The combination of eq.~\ref{eq:kw-classif} and~\ref{eq:kw-encoder} gives:
\begin{equation}
    \y_k = \sigma\left( E(\pi_k; \theta_E) \ast F(\x; \theta_F) \right).
\end{equation}
We want $y_{k, t} \approx 1$ when the phone sequence $\pi_k$ appears in the audio input $\x$ and ends at $t$, and $0$ otherwise. 
It is therefore possible to jointly train $F$ and $E$ from a generic speech training set $\mathcal{D} = \{(\x^{(i)}, \pi^{(i)})\}$ with the cross-entropy loss, without knowing what keywords will be used at inference. 

For each time step $t$ of each training sample $\x^{(i)}$, we create a set $\mathcal{K}_{i,t}^+$ of positive keyword examples (i.e. $\pi_k \in \mathcal{K}_{i,t}^+$ are subsequences of $\pi^{(i)}$ ending at time $t$), 
and a set of negative keyword examples $\mathcal{K}_{i,t}^-$,
and we minimize the following cross-entropy loss:
\begin{equation}
    \mathcal{L}_{KWS} =  \sum_{\x^{(i)}, \pi^{(i)} \in \mathcal{D}} \sum_{t} \ell(i, t),
\end{equation}
where
\begin{equation}
    \ell(i, t) = - \sum_{k \in \mathcal{K}_{i,t}^-} \log (1 - y^{(i)}_{k,t}) - \sum_{k \in \mathcal{K}_{i,t}^+} \log y^{(i)}_{k,t} \label{eq:kws-loss}.
\end{equation}

\subsection{Dataset creation}

In order to build the set $\mathcal{K}_{i,t}^+$ of positive keyword examples for dataset sample $i$ at frame $t$, we need to know what are the last phones appearing in the utterance at time $t$. They may be inferred from the forced alignment of the utterance with the ground-truth phone sequence.

To obtain them, we first train the stack of LSTMs on $\mathcal{D}$ to predict the phone sequences with the connectionist temporal classification loss (CTC~\cite{graves2006connectionist}) $\mathcal{L}_{CTC}$:
\begin{equation}
    \mathcal{L}_{CTC} = - \sum_{(\x, \pi) \in \mathcal{D}} \log \sum_{\l : \mathcal{B}(\l) = \pi} \prod_t p(l_t | \x), 
\end{equation}
where $\l = l_1 \ldots l_T$ is a sequence of phone or \textit{blank} labels and $\mathcal{B}$ is the CTC collapse function that removes label repetitions and \textit{blank} labels.
This network is used to align the dataset, i.e., for each $(\x, \pi) \in \mathcal{D}$, compute:
\begin{equation}
    \l^*(\x, \pi) = l^*_1 l^*_2 \ldots l^*_T = \arg\max_{\l : \mathcal{B}(\l) = \pi} \prod_t p(l_t | \x).
\end{equation}

Let $W$ be the receptive field of the convolutional network. By removing \textit{blanks} and label repetitions, $\mathcal{B}(l^*_{t-W} \ldots l^*_{t})$ will actually yield the sequence of phones in the utterance between time $t-W$ and $t$. Any suffix of that sequence may be considered an example of positive keyword to be detected at $t$. 
The set of positive examples $\mathcal{K}_t^+$ at time $t$ is created by randomly sampling suffixes of $\mathcal{B}(l^*_{t-W} \ldots l^*_{t})$ of length $3$ to $10$.
Note that the sampled ``keyword'' phone sequences do not necessarily correspond to actual words during training.

Since the network is trained with batches of dataset samples, the set of negative keyword examples $\mathcal{K}_t^-$ for one sample can merely be the union of sets of positive examples for the other samples in the batch. 

\subsection{Adaptation on specific dataset}
\label{sec:finetune}
Acoustic KWS approaches such as~\cite{bluche2020} rely on phone predictions and might not gain much from fine-tuning the model on a keyword-specific dataset.
In the proposed approach, the weights generated by the keyword encoder could serve as a starting point for re-training on keyword-specific data, when available. That would amount to rewrite eq.~\ref{eq:kw-encoder} as:
\begin{equation}
    \theta_k = E(\pi_k; \theta_E) + \theta^{(data)}_k.
\end{equation}
Given a training dataset $\mathcal{D}_{\mathcal{K}}$ for a set of keywords $\mathcal{K}$ made of positive example of keywords and negative data, the data-specific parameters $\{\theta^{(data)}_k\}_{k \in \mathcal{K}}$ can be adjusted by gradient descent, to optimize the loss of eq.~\ref{eq:kws-loss}.
By adjusting only those parameters, the ability of the model to detect any other arbitrary keyword is not lost.

\section{Experimental results}
\label{sec:results}

\subsection{Experimental setup}
\label{sec:setup}

We trained quantized LSTM networks on Librispeech~\cite{panayotov2015librispeech} with CTC and data augmentation. The inputs are MFCC and outputs are 46 phones plus a \textit{blank} class. The details of the training and quantization procedure can be found in ~\cite{bluche2020}. 
The weights of the LSTM cells are frozen after CTC training and not fine-tuned during the training of the KWS network, allowing to use them for ASR as well or with the method of~\cite{bluche2020}. We evaluated systems based on two such LSTM networks, with five layers of 64 and 96 units. 

The top softmax layer in that model is replaced by two convolutional layers to build the KWS network (shown in purple and green in Figure~\ref{fig:nf-model}). The first convolutional layer has a kernel of five frames, and 96 output tanh channels, followed by a max-pooling of size three and stride two. The top convolutional layer has a kernel size of 12. The total receptive field of the convolutional part has a size of $30$ frames. 
The keyword encoder is made of a bidirectional LSTM layer with 128 units in each direction, followed by a linear transform, predicting $12 \times 96 = 1152$ weights for each keyword. Overall, the model based on the \texttt{5x64} (\textit{resp.} \texttt{5x96}) LSTM have 208.8k (\textit{resp.} 440.7k) parameters plus 1.2k  parameters per keyword.

The keyword detector and encoder are jointly trained for five epochs using minibatches of 128 audio samples and two positive synthetic keyword samples in each $\mathcal{K}_{i}^+$, using the Adam optimizer and a learning rate of 1e-4, following the procedure presented in Section~\ref{sec:training}. 
All the weights, including those predicted by the keyword encoder are quantized to 8 bits.

\subsection{Keyword spotting results}

For this task, we crowd-sourced queries for two use-cases: a smart light scenario and a washing machine scenario\footnote{The datasets are publicly available at \url{https://bit.ly/39YV1te}} (cf.~\cite{bluche2020} for details). Each dataset was re-recorded in clean and noisy, reverberated far-field conditions with a SNR of 5dB. Each query contains between one and four keywords, and is expressed in natural language (e.g. ``could you \texttt{[turn on]} the lights in the \texttt{[bedroom]}").  
We measure the ratio of exactly parsed queries, i.e.~those for which the sequence of detected keywords exactly matches the expected one, 
and the F1 score as a measure of performance at the keyword level. 


The proposed KWS model is compared to five baselines, presented in detail in~\cite{bluche2020}. They all include the quantized LSTM acoustic model which was used as a base model for the KWS neural network. The \textit{Viterbi} and \textit{Lattice} baselines are LVCSR baselines, where the keywords are detected in the Viterbi decoding or in the decoding lattice of the utterance with a vocabulary of 200k and a trigram language model. The \textit{Filler} baseline employs a standard keyword-filler model, where the filler model is a phone loop. The \textit{Greedy} and \textit{Sequence} baselines correspond to the approach presented in~\cite{bluche2020}, similar to~\cite{hwang2015online}, with two post-processing methods.

\begin{figure}[!htb]
  \includegraphics[width=\linewidth]{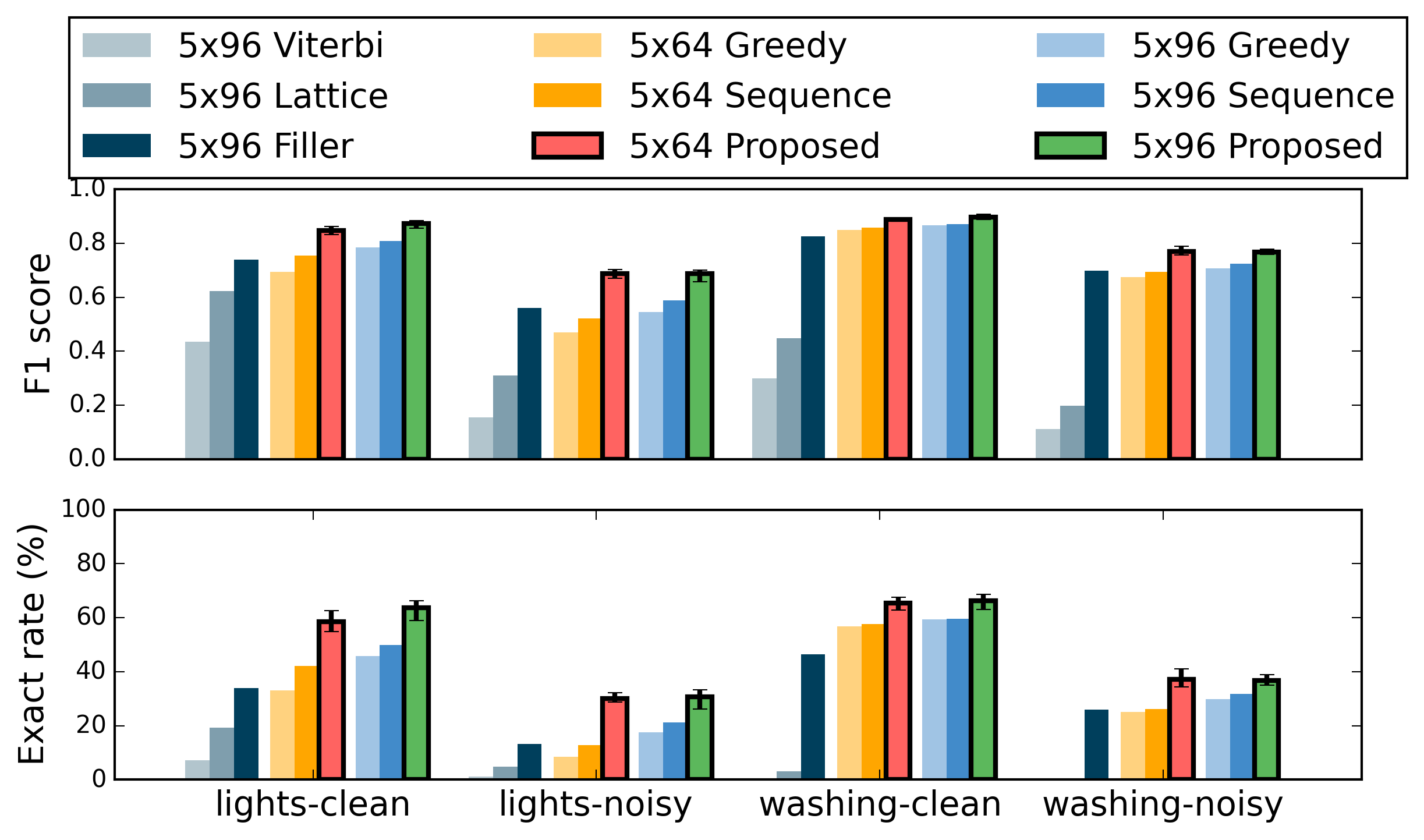}
\caption{F1 scores and exact rates of the proposed method (in red and green, showing the averaged results over 5 runs and the worst and best ones), compared to baselines on two keyword spotting tasks in clean and noisy conditions}
\label{fig:slu-results}
\end{figure}

\begin{table}[!tb]
\caption{F1 score for the proposed model and the \textit{Sequence} baseline from~\cite{bluche2020} on two keyword spotting tasks in clean and noisy conditions.\label{tab:slu-results}}
\centering
\begin{tabular}{lrrrrr}
\cline{2-6}
        & Dataset & \multicolumn{2}{c}{lights} & \multicolumn{2}{c}{washing} \\
$\downarrow$ Model    & $\longrightarrow$ & clean        & noisy       & clean        & noisy        \\ \hline
5x64 & \cite{bluche2020} & 0.754 & 0.522 & 0.857 & 0.694 \\
     & Proposed & 0.850 & 0.671 & 0.884 & 0.765 \\
     & \textit{(worst)} & \textit{0.835} & \textit{0.654} & \textit{0.878} & \textit{0.751} \\
     & \textit{(best)} & \textit{0.864} & \textit{0.687} & \textit{0.889} & \textit{0.784} \\\hline
5x96 & \cite{bluche2020} & 0.808 & 0.588 & 0.871 & 0.725 \\
& Proposed & \textbf{0.873} & \textbf{0.694} & \textbf{0.900} & \textbf{0.773} \\
& \textit{(worst)} & \textit{0.856} & \textit{0.665} & \textit{0.892} & \textit{0.764} \\
& \textit{(best)} & \textit{0.883} & \textit{0.709} & \textit{0.911} & \textit{0.782} \\\hline
\end{tabular}
\end{table}

We trained five models with different random initialization.
The results
are displayed in Figure~\ref{fig:slu-results}
and compared to the \textit{Sequence} post-processing approach of~\cite{bluche2020} in Table~\ref{tab:slu-results}. The LVCSR-based results are low, although using recognition lattices provides a big improvement over Viterbi decoding. The keyword-filler model is the best of the traditional methods. The baselines from~\cite{bluche2020} are competitive with the filler model. With the same model size, they are almost always better, with both the greedy and sequence post-processors. With a smaller model, the sequence post-processor yields better results than the filler model. The proposed approach outperforms all baselines, even with the small model, which has two times less parameters.

\subsection{Analysis of learned filters}

\begin{figure*}[!t]
\centering
    \begin{subfigure}{0.27\linewidth}
  \centering
  \includegraphics[width=\linewidth]{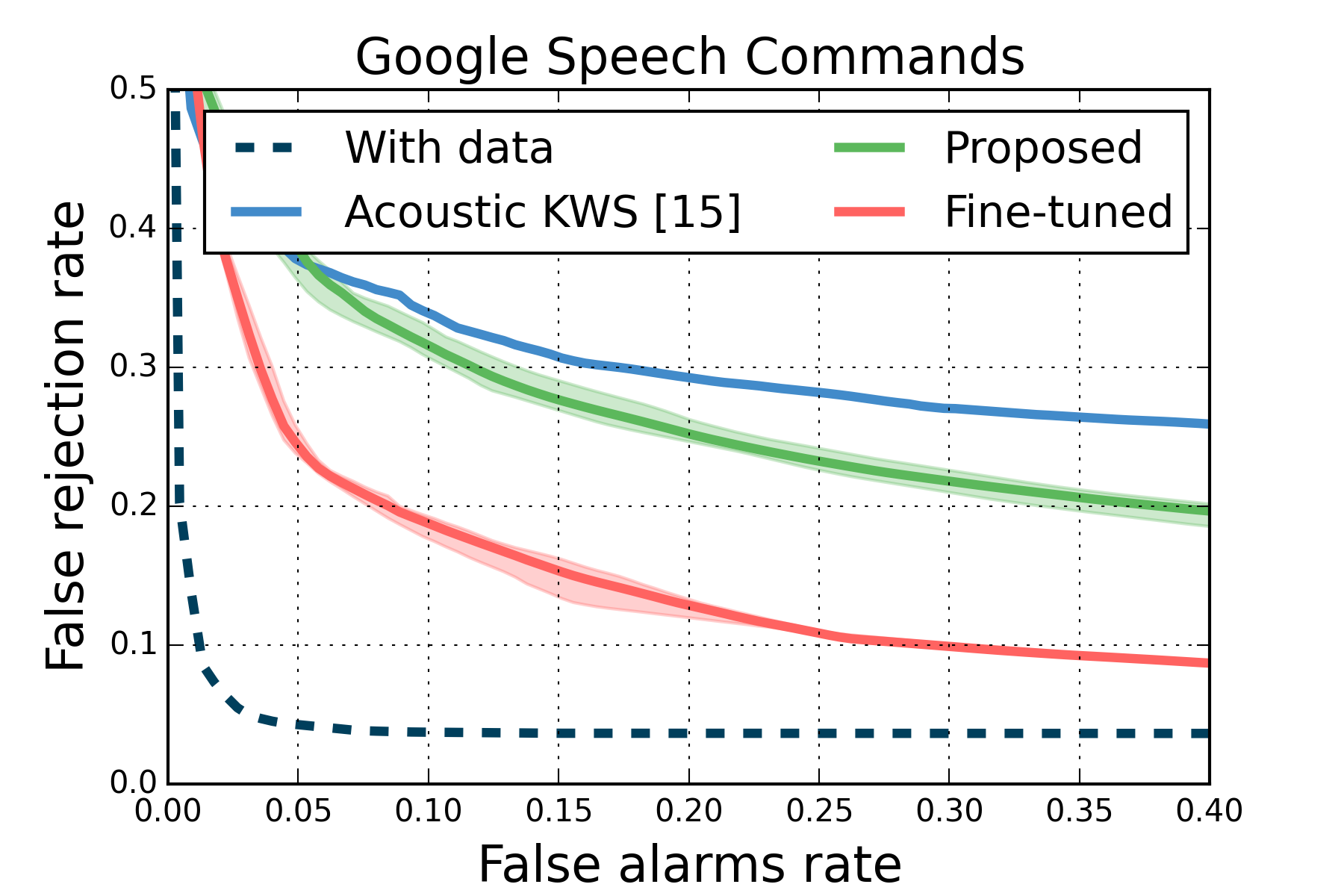}
  \caption{Google speech commands}
\end{subfigure}
    \begin{subfigure}{0.27\linewidth}
  \centering
  \includegraphics[width=\linewidth]{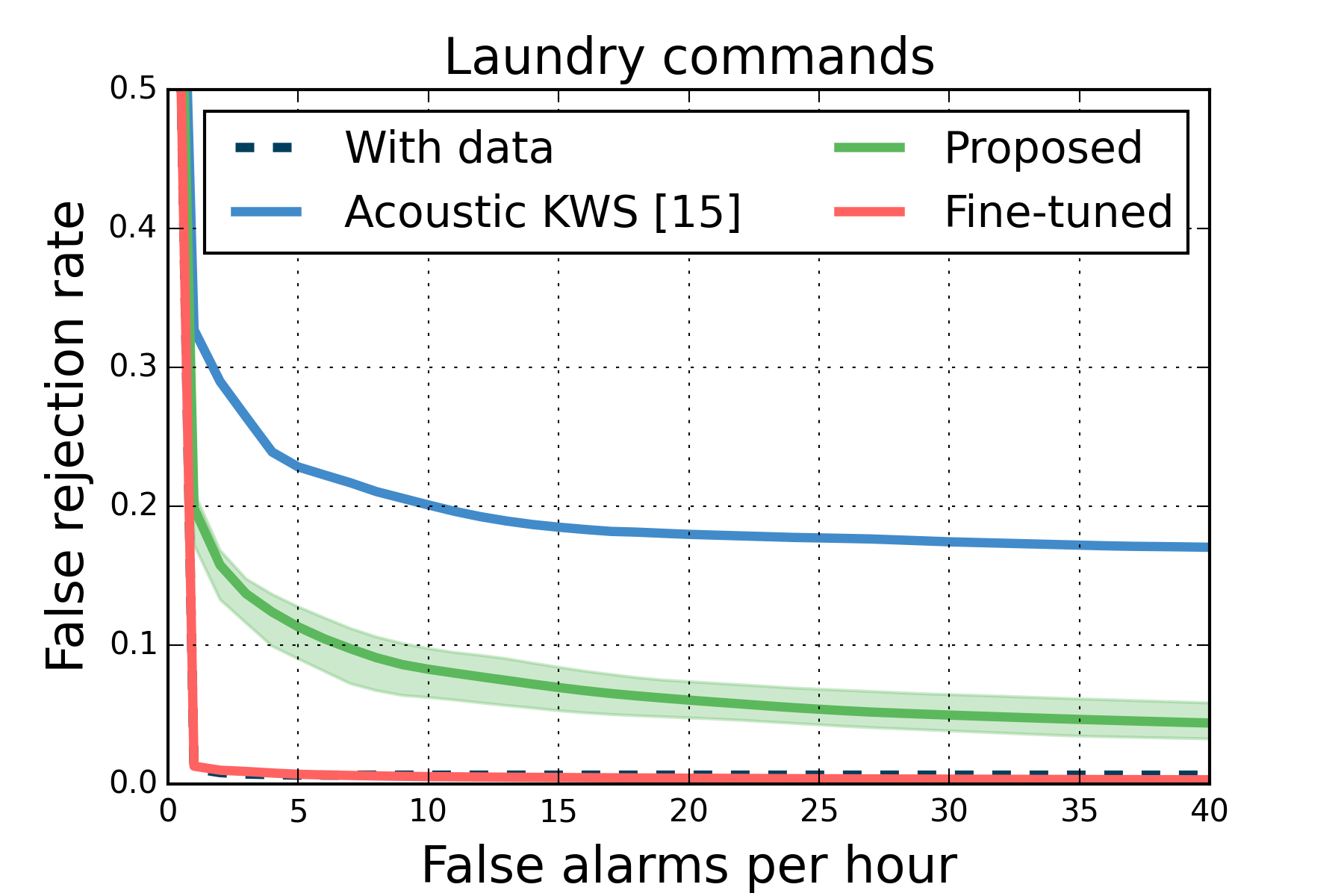}
  \caption{Laundry commands}
\end{subfigure}
\begin{subfigure}{0.27\linewidth}
  \centering
  \includegraphics[width=\linewidth]{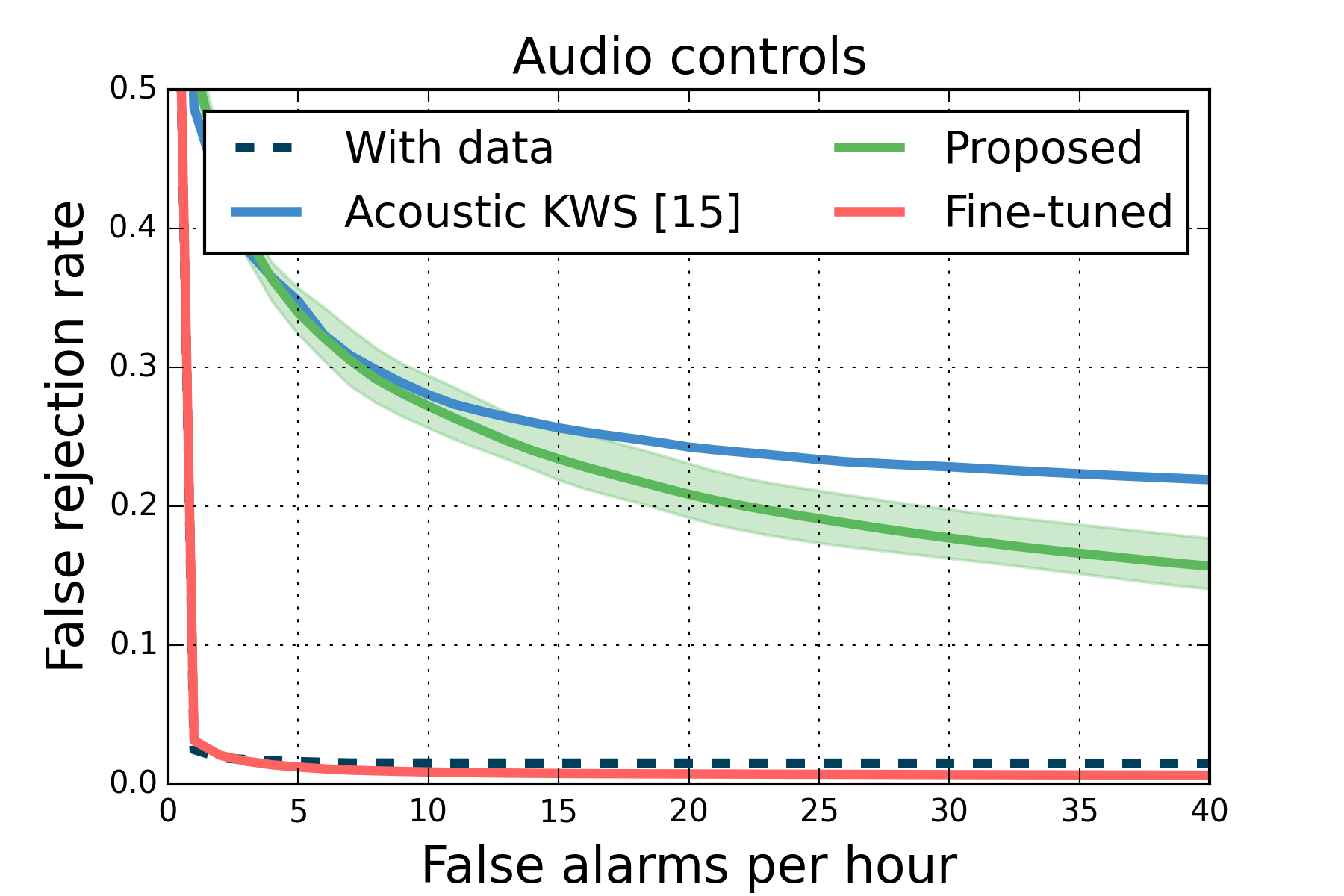}
  \caption{Audio controls}
\end{subfigure}
\caption{Speech commands results before (Section~\ref{sec:sc-results}, green) and after (Section~\ref{sec:finetune-res}, red) fine-tuning, averaged over 5 runs, compared with a classification model fully trained on specific training data (dashed black line) and to the acoustic KWS approach of~\cite{bluche2020} (blue). For Google speech commands, the standard evaluation measuring the false alarm rate on other commands is applied.}
\label{fig:sc-results}
\end{figure*}

To analyze what the keyword encoder has learned, we compare the predicted convolution filters for different inputs. In particular, we measure the Euclidean distance between predicted filters. Indeed, if the filters are close, the KWS network will tend to make similar predictions and potentially confuse the corresponding words.

\begin{table}[!htb]
\caption{Keywords and closest words in a vocabulary, measured as the Euclidean distance between the predicted filters.\label{tab:closest-filters}}
\centering
\begin{tabular}{rl}\hline
Keyword & Closest vocabulary words \\\hline
turn on & anon, non, turnin, fernand, maranon \\
decrease & crease, increase, encrease, greece \\
brightness & rightness, uprightness, triteness, greatness \\
bedroom & bathroom, begloom, broom, broome \\
play & flay, clay, blaye, splay, ley, lay \\
start & astarte, tart, stuart, upstart, sturt \\\hline
\end{tabular}
\end{table}

The filters for all the words in a vocabulary are computed and compared to the predicted filters for some keywords. In Table~\ref{tab:closest-filters}, we show the words with the closest filters to those of the keywords. We observe that they tend to be words with similar pronunciations, of about the same length or shorter, and mostly with the same or very similar suffixes. 

\begin{figure}[t]
\centering
  \includegraphics[width=0.6\linewidth]{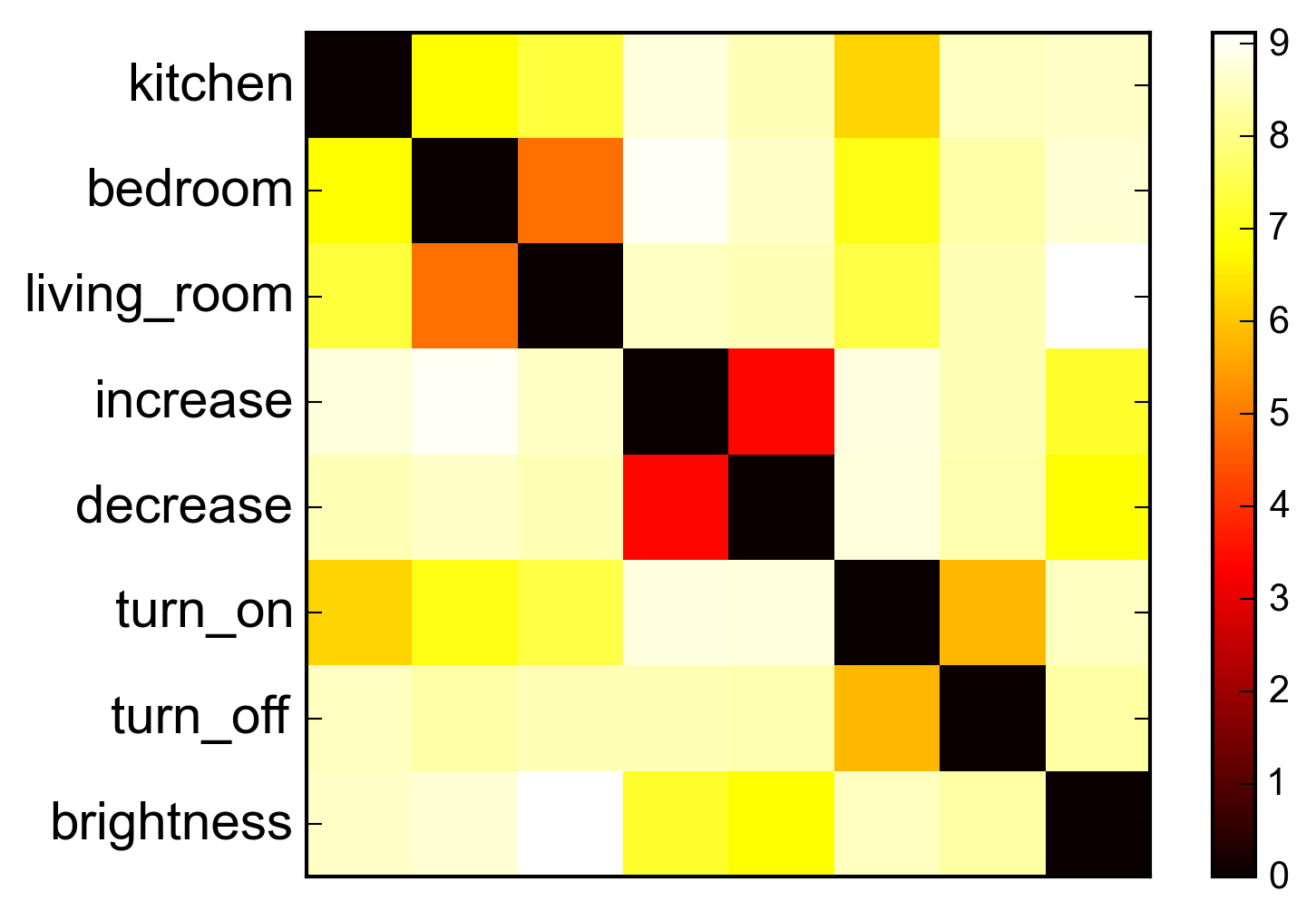}
\caption{Pairwise filter Euclidean distances between keywords of the \textit{lights} dataset (darker is closer).}
\label{fig:nf-filter-conf}
\end{figure}

Figure~\ref{fig:nf-filter-conf} shows the pairwise distances between the filters for the keywords in the set for the \textit{lights} dataset. We see that \texttt{increase} and \texttt{decrease} on the one hand, and \texttt{turn on} and \texttt{turn off} on the other hand are very close to one another, which correlates with the confusions of the KWS network: 37\% of the confusions are between  \texttt{increase} and \texttt{decrease}, 29\% for \texttt{turn on} and \texttt{turn off}. 


\subsection{Speech commands results}
\label{sec:sc-results}
For the speech command task, the goal is to detect a single command among a pre-defined set. We evaluate our approach 
on Google's speech command dataset~\cite{warden2018speech}\footnote{evaluating on the same 12 commands as in~\cite{sainath2015convolutional,tang2018deep}} and
on two in-house crowd-sourced datasets of speech commands: a
dataset of audio control commands with 10 commands (``turn on, turn off, play, pause, start, stop, next track, previous track, volume up, volume down'') and a dataset of laundry commands with 5 commands (``cancel, wash whites, wash delicate, wash heavy duty, wash normal''). We measure the false rejection rate across all commands on the command datasets and the number of false alarms per hour on a dataset of 45 hours of negative data made of music and speech. For the Google speech command dataset, we follow the usual evaluation procedure and measure the false alarm rate using the other commands as negative data.

We compare our results with the 5x64 model to a model trained on specific training data, similar in size (186K parameters) and performance to the \texttt{res15} ResNet architecture of~\cite{tang2018deep}, and to the \textit{Greedy} acoustic KWS baseline using the same 5x64 base network~\cite{bluche2020}. 
The results are depicted in Figure~\ref{fig:sc-results}. As expected, the model trained on specific training data (``with data'' in black) is much better than the other two open-vocabulary approaches. Nonetheless, the proposed method outperforms the acoustic KWS baseline. The results on Google speech commands look worse, but that dataset mainly consists of short commands of two or three phonemes, which are harder to discriminate, and the negative data in this case only contains commands too.

Moreover, the proposed model is mostly similar to the one used in the acoustic KWS baseline, since only the top layers are retrained, so they could be used jointly with low computation overhead.
Finally, the models labeled ``with data'' are models trained on specific training data 
while the proposed model can readily be applied to any keyword set without retraining. 

\subsection{Fine-tuning on keyword-specific training data}
\label{sec:finetune-res}
Since training data is available for these datasets, we fine-tuned the filters with the method presented in Section~\ref{sec:finetune}. The results are also shown (in red) in Figure~\ref{fig:sc-results}.
%
%
We see that after fine-tuning, the performance of the proposed model is similar to that of the model trained exclusively on specific training data. The results on Google speech commands are not as close: it might also be due to the fact that positive and negative data are short keywords in this dataset.

In these speech command tasks, the keywords are isolated: they are not inside a sentence. Before fine-tuning, the network was only trained on sentences. It is therefore possible that the gap between the network without and with fine-tuning (and ``with data'') is merely due to learning to detect silences surrounding the commands.  This should be explored in future work.
It is also worth noting that the fine-tuned network has not lost its ability to detect any arbitrary keyword, since only the weights of the top layer are modified. New keywords may then be added to the network without having to retrain it all. The baseline network does not offer these possibilities.

\section{Conclusion}
\label{sec:conclu}

We presented an open-vocabulary keyword spotting system, which does not require training data specific to the keywords to be detected at inference. In contrast to most acoustic keyword spotting models, it directly predicts a confidence score at the keyword-level, alleviating the need of a confidence calibration. We have shown that the proposed model outperforms acoustic KWS baselines for the detection of keyword both inside utterances and as isolated speech commands. We proposed a method to fine-tune the model to specific training data, which makes it as good as a speech command detector trained on specific data while retaining its ability to detect other arbitrary keywords.

\bibliographystyle{IEEEtran}
\bibliography{mybib}

\begin{thebibliography}{10}
\providecommand{\url}[1]{#1}
\csname url@samestyle\endcsname
\providecommand{\newblock}{\relax}
\providecommand{\bibinfo}[2]{#2}
\providecommand{\BIBentrySTDinterwordspacing}{\spaceskip=0pt\relax}
\providecommand{\BIBentryALTinterwordstretchfactor}{4}
\providecommand{\BIBentryALTinterwordspacing}{\spaceskip=\fontdimen2\font plus
\BIBentryALTinterwordstretchfactor\fontdimen3\font minus
  \fontdimen4\font\relax}
\providecommand{\BIBforeignlanguage}[2]{{%
\expandafter\ifx\csname l@#1\endcsname\relax
\typeout{** WARNING: IEEEtran.bst: No hyphenation pattern has been}%
\typeout{** loaded for the language `#1'. Using the pattern for}%
\typeout{** the default language instead.}%
\else
\language=\csname l@#1\endcsname
\fi
#2}}
\providecommand{\BIBdecl}{\relax}
\BIBdecl

\bibitem{xiong2016achieving}
W.~Xiong, J.~Droppo, X.~Huang, F.~Seide, M.~Seltzer, A.~Stolcke, D.~Yu, and
  G.~Zweig, ``Achieving human parity in conversational speech recognition,''
  \emph{IEEE/ACM Transactions on Audio, Speech, and Language Processing},
  p.~99, 2016.

\bibitem{saade2018spoken}
A.~Saade, A.~Coucke, A.~Caulier, J.~Dureau, A.~Ball, T.~Bluche, D.~Leroy,
  C.~Doumouro, T.~Gisselbrecht, F.~Caltagirone \emph{et~al.}, ``Spoken language
  understanding on the edge,'' \emph{NeurIPS Workshop on Energy Efficient
  Machine Learning and Cognitive Computing}, 2019.

\bibitem{helloedge}
\BIBentryALTinterwordspacing
Y.~Zhang, N.~Suda, L.~Lai, and V.~Chandra, ``Hello edge: Keyword spotting on
  microcontrollers,'' \emph{CoRR}, vol. abs/1711.07128, 2017. [Online].
  Available: \url{http://arxiv.org/abs/1711.07128}
\BIBentrySTDinterwordspacing

\bibitem{sainath2015convolutional}
T.~Sainath and C.~Parada, ``Convolutional neural networks for small-footprint
  keyword spotting,'' in \emph{Sixteenth Annual Conference of the International
  Speech Communication Association}, 2015.

\bibitem{coucke2019efficient}
A.~Coucke, M.~Chlieh, T.~Gisselbrecht, D.~Leroy, M.~Poumeyrol, and T.~Lavril,
  ``Efficient keyword spotting using dilated convolutions and gating,'' in
  \emph{ICASSP 2019-2019 IEEE International Conference on Acoustics, Speech and
  Signal Processing (ICASSP)}.\hskip 1em plus 0.5em minus 0.4em\relax IEEE,
  2019, pp. 6351--6355.

\bibitem{garofolo2000trec}
J.~S. Garofolo, C.~G. Auzanne, and E.~M. Voorhees, ``The trec spoken document
  retrieval track: A success story.'' \emph{NIST SPECIAL PUBLICATION SP}, vol.
  500, no. 246, pp. 107--130, 2000.

\bibitem{mamou2006spoken}
J.~Mamou, D.~Carmel, and R.~Hoory, ``Spoken document retrieval from call-center
  conversations,'' in \emph{Proceedings of the 29th annual international ACM
  SIGIR conference on Research and development in information retrieval}, 2006,
  pp. 51--58.

\bibitem{miller2007rapid}
D.~R. Miller, M.~Kleber, C.-L. Kao, O.~Kimball, T.~Colthurst, S.~A. Lowe, R.~M.
  Schwartz, and H.~Gish, ``Rapid and accurate spoken term detection,'' in
  \emph{Eighth Annual Conference of the international speech communication
  association}, 2007.

\bibitem{brown1997open}
M.~G. Brown, J.~T. Foote, G.~J. Jones, K.~S. Jones, and S.~J. Young,
  ``Open-vocabulary speech indexing for voice and video mail retrieval,'' in
  \emph{Proceedings of the fourth ACM international conference on Multimedia},
  1997, pp. 307--316.

\bibitem{chen2017confidence}
Z.~Chen, Y.~Zhuang, and K.~Yu, ``Confidence measures for ctc-based phone
  synchronous decoding,'' in \emph{2017 IEEE International Conference on
  Acoustics, Speech and Signal Processing (ICASSP)}.\hskip 1em plus 0.5em minus
  0.4em\relax IEEE, 2017, pp. 4850--4854.

\bibitem{szoke2005comparison}
I.~Szoke, P.~Schwarz, P.~Matejka, L.~Burget, M.~Karafi{\'a}t, M.~Fapso, and
  J.~Cernocky, ``Comparison of keyword spotting approaches for informal
  continuous speech,'' in \emph{Ninth European conference on speech
  communication and technology}, 2005.

\bibitem{hwang2015online}
K.~Hwang, M.~Lee, and W.~Sung, ``Online keyword spotting with a character-level
  recurrent neural network,'' \emph{arXiv preprint arXiv:1512.08903}, 2015.

\bibitem{lengerich2016end}
C.~Lengerich and A.~Hannun, ``An end-to-end architecture for keyword spotting
  and voice activity detection,'' \emph{arXiv preprint arXiv:1611.09405}, 2016.

\bibitem{chen2018sequence}
Z.~Chen, Y.~Qian, and K.~Yu, ``Sequence discriminative training for deep
  learning based acoustic keyword spotting,'' \emph{Speech Communication}, vol.
  102, pp. 100--111, 2018.

\bibitem{bluche2020}
T.~Bluche, M.~Primet, and T.~Gisselbrecht, ``Small-footprint open-vocabulary
  keyword spotting with quantized lstm networks,'' \emph{arXiv preprint
  arXiv:2002.10851}, 2020.

\bibitem{lugosch2018donut}
L.~Lugosch, S.~Myer, and V.~S. Tomar, ``Donut: Ctc-based query-by-example
  keyword spotting,'' \emph{arXiv preprint arXiv:1811.10736}, 2018.

\bibitem{yang2018automatic}
Y.~Yang, A.~Lalitha, J.~Lee, and C.~Lott, ``Automatic grammar augmentation for
  robust voice command recognition,'' \emph{arXiv preprint arXiv:1811.06096},
  2018.

\bibitem{audhkhasi2017end}
K.~Audhkhasi, A.~Rosenberg, A.~Sethy, B.~Ramabhadran, and B.~Kingsbury,
  ``End-to-end asr-free keyword search from speech,'' \emph{IEEE Journal of
  Selected Topics in Signal Processing}, vol.~11, no.~8, pp. 1351--1359, 2017.

\bibitem{sacchi2019open}
N.~Sacchi, A.~Nanchen, M.~Jaggi, and M.~Cernak, ``Open-vocabulary keyword
  spotting with audio and text embeddings,'' in \emph{INTERSPEECH 2019-IEEE
  International Conference on Acoustics, Speech, and Signal Processing}, no.
  CONF, 2019.

\bibitem{klein2015dynamic}
B.~Klein, L.~Wolf, and Y.~Afek, ``A dynamic convolutional layer for short range
  weather prediction,'' in \emph{Proceedings of the IEEE Conference on Computer
  Vision and Pattern Recognition}, 2015, pp. 4840--4848.

\bibitem{noh2016image}
H.~Noh, P.~Hongsuck~Seo, and B.~Han, ``Image question answering using
  convolutional neural network with dynamic parameter prediction,'' in
  \emph{Proceedings of the IEEE conference on computer vision and pattern
  recognition}, 2016, pp. 30--38.

\bibitem{jia2016dynamic}
X.~Jia, B.~De~Brabandere, T.~Tuytelaars, and L.~V. Gool, ``Dynamic filter
  networks,'' in \emph{Advances in Neural Information Processing Systems},
  2016, pp. 667--675.

\bibitem{graves2006connectionist}
A.~Graves, S.~Fern{a}ndez, F.~Gomez, and J.~Schmidhuber, ``Connectionist
  temporal classification: labelling unsegmented sequence data with recurrent
  neural networks,'' in \emph{Proceedings of the 23rd international conference
  on Machine learning}.\hskip 1em plus 0.5em minus 0.4em\relax ACM, 2006, pp.
  369--376.

\bibitem{panayotov2015librispeech}
V.~Panayotov, G.~Chen, D.~Povey, and S.~Khudanpur, ``Librispeech: an asr corpus
  based on public domain audio books,'' in \emph{2015 IEEE International
  Conference on Acoustics, Speech and Signal Processing (ICASSP)}.\hskip 1em
  plus 0.5em minus 0.4em\relax IEEE, 2015, pp. 5206--5210.

\bibitem{warden2018speech}
P.~Warden, ``Speech commands: A dataset for limited-vocabulary speech
  recognition,'' \emph{arXiv preprint arXiv:1804.03209}, 2018.

\bibitem{tang2018deep}
R.~Tang and J.~Lin, ``Deep residual learning for small-footprint keyword
  spotting,'' in \emph{2018 IEEE International Conference on Acoustics, Speech
  and Signal Processing (ICASSP)}.\hskip 1em plus 0.5em minus 0.4em\relax IEEE,
  2018, pp. 5484--5488.

\end{thebibliography}
\end{document}